This is an author manuscript, published in
Anne Wagner (ed), *International Handbook of Legal Language and Communication* (Springer Nature 2026).
https://doi.org/10.1007/978-3-031-87993-7_100-1 please cite the VoR only

# Legal Interpretation and AI: From Expert Systems to Argumentation and LLMs

Václav Janeček* and Giovanni Sartor†

**Abstract**

AI and law research has encountered legal interpretation in different ways, in the context of its evolving approaches and methodologies. Research on expert systems has focused on legal knowledge engineering, with the goal of ensuring that human-generated interpretations can be precisely transferred into knowledge bases, to be consistently applied. Research on argumentation has aimed at representing the structure of interpretive arguments, as well as their dialectical interactions, to assess the acceptability of interpretive claims within argumentation frameworks. Research on machine learning has focused on automated generation of interpretive suggestions and arguments, through general and specialized generative AI tools, now being increasingly deployed in legal practice.

Keywords: Interpretation, Artificial intelligence, Expert systems, Argumentation, Large language models

## 1. Introduction

This chapter discusses how AI and law research tackles legal interpretation. We first introduce the concept of legal interpretation (Section "The Concept of Legal Interpretation") and then move into the ways in which interpretation has been addressed within different strands of AI and law research: legal expert systems (Section "Legal Interpretation in Expert Systems"), argumentation (Section "Legal Interpretation in Argumentation"), and finally machine learning and generative AI (Section "Interpretation and Machine Learning (LLMs)").

## 2. The Concept of Legal Interpretation

Legal interpretation is a pervasive aspect of legal practice, at all levels and for all legal professions: judges, lawyers, public administrators, and legislators. It is also a task which concerns all citizens, whenever they have to figure out what the law requires of them or how it can help them. A vast literature exists on legal interpretation, adopting various theoretical approaches, typically focusing on specific legal systems or sections of it. Here we shall not enter the jurisprudential debate on interpretation, but just shortly introduce the concept of interpretation in order to specify the focus of this chapter.

---

* University of Bristol Law School; IGSG-CNR (Leverhulme International Fellow). V. Janeček's research was funded by The Leverhulme Trust (Leverhulme International Fellowship: Probabilistic Legal Interpretation).

† University of Bologna; European University Institute. G. Sartor's research was funded by the European Research Council (ERC) under the European Union's Horizon 2020 research and innovation programme (Grant Agreement No. 833647: CompuLaw).

The concept of legal interpretation, broadly understood, covers every attempt at determining the meaning of a legal text (most typically a statutory provision) and every outcome of such interpretive attempts. Importantly, legal interpretation often seeks to determine the meaning of a provision with a view to apply it (or not) to a specific scenario and eventually to act on the rule contained in the interpreted provision.

Often, a more restrictive concept of interpretation is adopted, focusing only on those attempts at determining the meaning which are motivated by an initial stage of doubt. If there is no doubt about the precise meaning of the interpreted provision, the legal rule can be applied based on an automatic and unreflected “understanding” of the text, without the need for any interpretation. This idea is expressed by the traditional saying *in claris non fit interpretatio* (Dascal and Wróblewski 1988). Doubts may arise from an unprecise formulation of the text as well as from various contextual-pragmatic conditions, e.g., the fact that the prima facie meaning would lead to undesirable consequences, or that it would lead to implications not fitting the plausible intentions of legislators or not matching important social values or widespread conditions. Consider, for instance, whether the rule “No vehicles in the park” (Hart 1961) covers bicycles or scooters ridden or driven by children or adults, and what undesirable consequences might spring from different interpretations of that provision.

The outcomes of interpretative attempts can be of two main kinds: (*a*) declarative interpretations that ascribe to the interpreted provision the meaning that is the most immediate and plausible according to semantic rules and generally shared pragmatic presuppositions, (*b*) interpretations that deviate from declarative meanings by broadening (extensive interpretations) or narrowing (restrictive interpretations) the scope of a provision. Thus, going back to the above example, an interpretation of “vehicle” that also applies to e-scooters may be viewed as declarative, an interpretation that extends it to kick-scooters may look extensive (assuming vehicles are usually assumed to be motorized), and an interpretation that excludes e-scooters may look rather restrictive. Extensive interpretation comes close to analogy (Canale and Tuzet 2017) which however is beyond the scope of this chapter.

# 3. Legal Interpretation in Expert Systems

In its initial stage (1970s–1990s), AI and law research focused on the so-called symbolic AI, based on human-generated knowledge representation. It was generally assumed that to develop an intelligent system, humans had to provide a formal representation of the relevant knowledge (usually expressed through a combination of rules and concepts), coupled with algorithms for making inferences out of such knowledge. Different logical formalisms (rule languages, classical logic, modal and descriptive logics, formal argumentation, etc.) and computable models for inferential processes (deductive, defeasible, inductive, probabilistic, case-based, etc.) were developed and applied.

The typical structure of a system based on symbolic knowledge is represented in Fig. 1. Note that humans appear both as users of the system and as creators of the system’s knowledge base (experts, possibly helped by knowledge engineers).

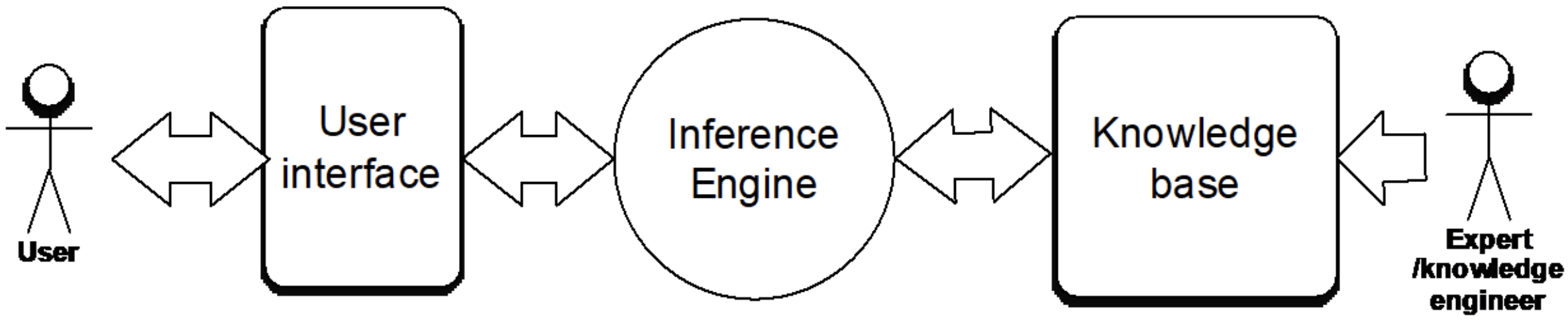

*Fig. 1 Knowledge-based system*

In the context of research on knowledge-based systems, the issue of legal interpretation concerns mainly the human knowledge engineer and users of the systems: it asked how legal sources should be modelled so that the correct legal interpretation is transferred into the knowledge-base of the system, to be consistently applied to the cases submitted to the system.

The relation between knowledge representation and the solution of interpretive doubts emerged in the context of the research by Layman Allen and Charles Saxon, who emphasized doubts emerging from syntactic ambiguities and the possibilities to rigorously express interpretive choices by injecting logical structures into legal texts through logical connectives and formatting (Allen and Saxon 1995). Consider, for instance, the statement: "Individuals with a degree in law and in computing may apply." This statement is syntactically ambiguous because it is not clear whether only individuals having both degrees may apply, or whether one of the two degrees is sufficient. Allen and Saxon argued that the drafter (or the knowledge-based engineer) should use formal representations such as:

IF An individual has a degree in law OR
The individual has a degree in computing
THEN The individual may participate.

This formulation clarifies that one degree is sufficient, eliminating the syntactic ambiguity (if the interpretation is that both degrees are needed, the logical connective "OR" should be substituted with an "AND"). The analysis by Allen and Saxon showed that the creation of a knowledge-based system involved interpretive choices with regard to doubts arising from syntactic ambiguities, which should be solved in order to select the interpretation to be included in the knowledge base.

Further studies also addressed interpretive doubts resulting from ambiguity/vagueness pertaining to the meaning of legal texts that are being modelled in a knowledge base. In the context of a knowledge-based system, the uncertainty around the meaning of the text and the terms used therein can be addressed in two basic ways: (*a*) at the time when the system is applied to a specific case, by the officer tasked with that case or (*b*) at the time when the knowledge base of the system is prepared, by the knowledge engineer tasked with the drafting or revision of the knowledge base.

In the first case, the system will ask the officer whether the term to be interpreted applies to the case (the system may additionally support the officer by providing access to relevant documents such as legal sources, doctrines, etc.). In the second case, the knowledge base is expanded with further rules and concepts meant to specify the chosen interpretation (Sergot et al. 1986).

Consider, for instance, the abovementioned interpretive doubts concerning the "No vehicles in the park" provision. Assume that the knowledge-based system contains the following rule:

Rule 1*: x* is prohibited in the park IF x is a vehicle.

Assume that John would like to enter the park with his device, which is a racing bike. To positively answer the query "Is it the case that John's device is prohibited in the park?", the system needs the information that this device is a vehicle.

Following the first approach, the information would be provided to the system directly by John, who may be queried by the system: "Is it the case that your device is a vehicle?" If John says "Yes", this information would be added to the system's knowledge base, so that the system could conclude, by applying the Rule 1, that "Your device is prohibited in the park!".

Following the second approach, the knowledge engineer could ex ante expand the system's knowledge base with an interpretive rule such as the following:

Rule 2: x is a vehicle IF x is bicycle.

Based on this additional rule, the system will ask John: "Is it the case that your device is a bicycle?". Assume that John gives a positive answer. Then, by chaining Rule 1 and Rule 2, the system will be able to conclude that John's device is prohibited from the park. Under this second approach, the task of determining whether a bicycle is a vehicle has been taken over by the knowledge engineer.

The expansion of the knowledge-based system with interpretive rules (such as Rule 2 above) raised an issue concerning the allocation of the "interpretive power": to the extent that interpretive rules are embedded in the knowledge base, the task of interpretation is transferred from the officer responsible for applying the rule in concrete situations to the knowledge engineer who builds the knowledge-based system before it is deployed. This more abstract (and ex-ante) approach to interpretation has been criticized for not being sensible to specific circumstances. For instance, it may be argued that while the officer tasked with applying Rule 1 could have easily concluded that the children's bike of little Tom does not count as a vehicle, it would have been more difficult for her to arrive at that child-friendly conclusion had the knowledge engineer already introduced in the knowledge base the interpretive rule stating that bikes are vehicles. This is because once an overbroad interpretive rule is introduced in the system, it becomes difficult to put that interpretive rule aside without there being an additional exception-granting mechanism in the AI system.

# 4. Legal Interpretation in Argumentation

The knowledge-based approach relegated interpretation to the margins of AI and law research, entrusting it to the users of the system or to the knowledge engineers who supplement the system with the additional interpretive rules. This was because a logically consistent knowledge base—to be applied through deductive reasoning—can include provisions and terms that are susceptible to different interpretations, but it cannot include alternative incompatible interpretations of such provisions and terms.

And so without a robust approach for resolving interpretive inconsistencies at the system's level, this task was kept within the purview of humans.

Eventually, this limitation of AI systems in respect of legal interpretation was overcome by AI and law approaches focusing on argumentation, namely, on the explicit opposition of opposite points of view, as expressed in competing arguments. Following this approach a knowledge base does not need to be logically consistent, since the system can build and assess arguments based on conflicting premises extracted from the knowledge base. In the context of argumentation, interpretation can thus be modelled through the dialectical interaction of alternative proposals for meaning ascription, each based on corresponding interpretive canons. The canons being considered may include, for instance, interpretation from (1) ordinary meaning, (2) technical meaning, (3) contextual harmonization, (4) precedent, (5) statutory analogy, (6) a legal concept, (7) general principle, (8) history, (9) purpose, (10) substantive reasons, or (11) intention (MacCormick and Summers 1991). Methodologically, the canonical approaches to legal interpretation may vary a little across different jurisdictions (Gábriš and Žák Krzyžanková 2025), but, for example, the ordinary language canon (Slocum 2015) is recognized universally.

Here is one formulation of the ordinary language interpretive canon:

> IF interpreting *E* as including (OR excluding) *S* fits Ordinary Language, THEN *E* should be interpreted as including (OR excluding) *S*.

Note that interpretive canons, as the other conditionals in our examples, are defeasible conditionals (Sartor 2012): their antecedents provide a sufficient ground for endorsing the consequent, but only as long as there are no prevailing reasons to the contrary. Such canons may be used for canonical expressions of meaning ascription, such as the claim that a term should be interpreted as meaning, excluding or including a certain content (Sartor et al. 2014; Walton et al. 2018, 2021), while the outcomes of such rules can be modelled through constitutive rules ascribing legal qualifications according to that content (Boella et al. 2010).

Let us consider now the case of little Tom, who would like to enter the park with the small bicycle he received for Christmas. The park guardian could stop him by arguing that the term "vehicle" in the rule "No vehicles in the park" includes bicycles, since, in ordinary language the term vehicle covers any transportation means that can carry one or more persons. This argument could be formalized by combining a specification of the ordinary language canon

1. IF interpreting "*vehicle*" as including *bicycles* fits Ordinary Language, THEN "*vehicle*" should be interpreted as including *bicycles*,

with the assertion that

2. interpreting *"vehicle"* as including *bicycles* fits the Ordinary Language,

to achieve the conclusion that indeed

3. "*vehicle*" should be interpreted as including *bicycle*s.

Note that premise (2) can be supported by further arguments (omitted here), e.g., by some factual evidence that the term "vehicle," as ordinarily understood, covers any kind of transportation means for people.

The officer's argument can be countered by little Tom, for instance, on the basis of an instantiation of canonical formulation of the teleology canon (Purposiveness, see Canale and Tuzet 2023):

1. IF interpreting "vehicle" as excluding *children's bicycles* fits Purposiveness, THEN "*vehicle*" should be interpreted as excluding *children's bicycles*,

and the assertion that

2. interpreting "*vehicle*" as excluding *children's bicycles* fits Purposiveness,

leading to the conclusion that

3. "*vehicle*" should be interpreted as excluding *children's bicycles*.

Note that premise (2) of Tom's counter-argument can be supported by a further argument, based for instance on the fact that interpreting "vehicle" as excluding children's bicycles fits the purposes of the law since it facilitates the enjoyment of the park to families and has no significant negative impacts on the peaceful and safe enjoyment of it by others. Of course, it is an exaggeration to assume that little Tom would make this formal argument. Rather, Tom would flag the officer's legal conclusion as being unfair. But the example shows that the model argument can formally justify the link between one's interpretive views about a provision and the legal conclusion that stems from it.

The bridge between interpretative statements and substantive legal conclusions can be based on the idea that interpretative statements ground constitutive rules, namely, rules attributing classifications (Boella et al. 2010):

> IF *E* should be interpreted as including (excluding) *S,* THEN [IF *X* is a *S, THEN X* is (NOT) an *E*].

An instance would be the following:

> If "*vehicle*" should be interpreted as including *bicycles,* THEN (IF *Tom's device* is a bicycle, THEN *Tom's device* is a *vehicle*).

All argument steps mentioned so far are visualized in the argument graph in Fig. 2, where each conclusion (final or intermediate) is connected by lines to the premises supporting it (where the letter P in the linked oval denotes that this is a presumptive inference). The arrow denotes attacks between arguments having incompatible conclusions.

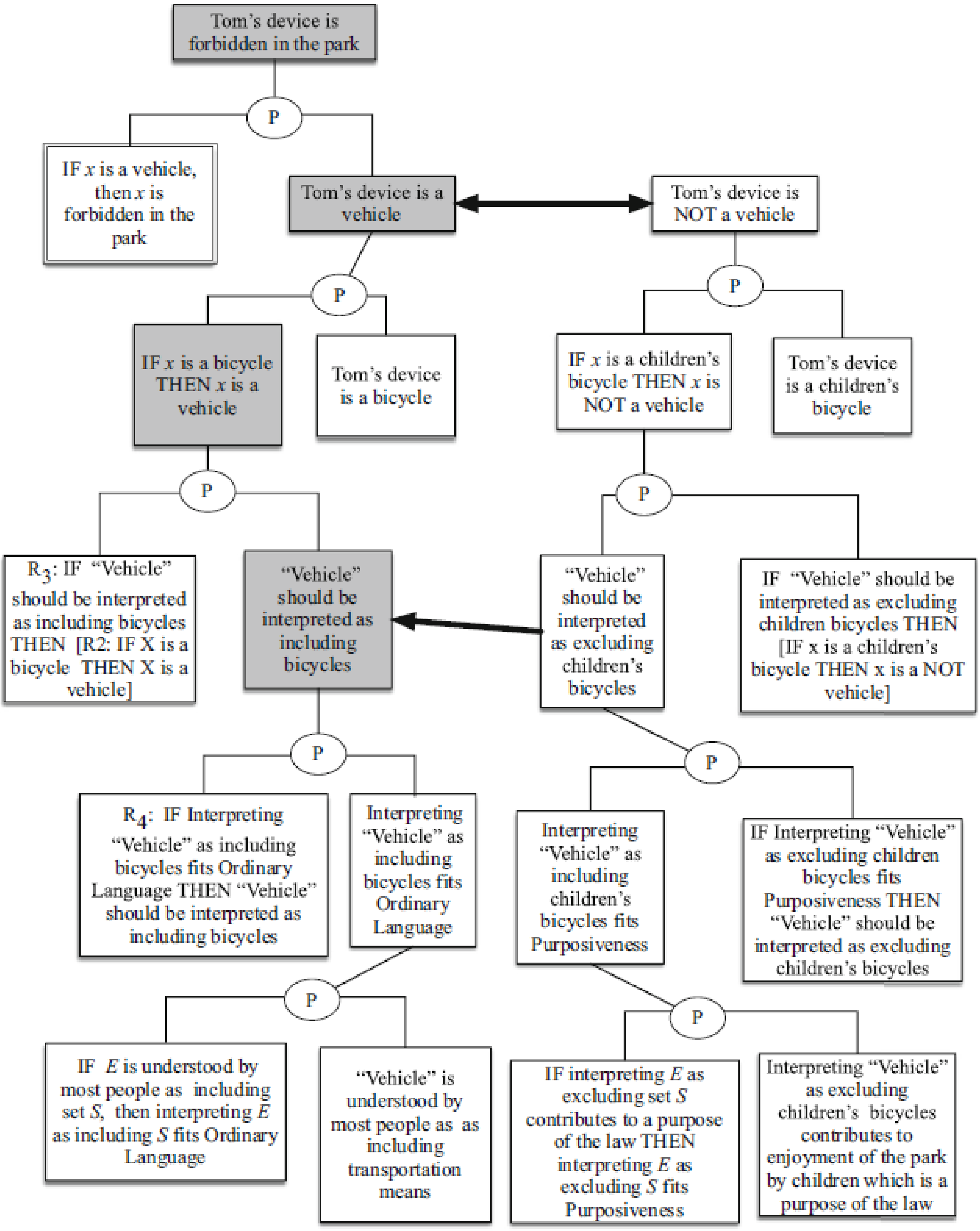

*Fig. 2 Interpretive arguments*

The analysis of the graph can be accomplished according to a semantic for formal argumentation, according to which arguments are “IN” (i.e., they are exempt from relevant criticisms, in the context of the graph) if they are not attacked by arguments that are “IN” (for details, see Sartor 2023). According to this approach, the whole argument to the left is “OUT,” since it is successfully attacked by the interpretive argument to the right (the one based on the purpose of the rule, according to the canon of Purposiveness), which is considered to be stronger. Note that in this model for (structured) argumentation we assume that an argument can only attack arguments that are not stronger than itself (see Prakken 2010). Note also that where a premise is “OUT,” all consequences depending on that premise are also “OUT,” as is indicated by the gray boxes in Fig. 2. Thus, according to the graph, the officer would fail to establish that Tom’s device is a vehicle, since her argument to that effect, based on ordinary language, is defeated by the prevailing teleological argument that excludes that Tom’s device counts as a vehicle.

Argumentation-based approaches may also include arguments to the effect that one argument is stronger than other arguments, for instance, to the effect that teleological arguments usually prevail over ordinary language arguments (for a discussion, see Sartor 2023). In AI and law, further work has been developed on argumentation-based analyses of interpretation adopting different formalisms, e.g., defeasible logic (Malerba et al. 2021) or fuzzy logic (da Costa Pereira et al. 2017).

# 5. Interpretation and Machine Learning (LLMs)

While some attempts have been done to model legal concepts through neural networks (Philipps and Sartor 1999), in particular in domains where no precise rules are available (Zeleznikow and Stranieri 1995; Ruggeri et al. 2022), the main expectations and results for the use of machine learning in legal interpretation pertain today to generative AI and particularly to large language models.

Generative AI (GenAI) includes a family of increasingly complex AI systems endowed with the ability to generate high-quality content. Here we focus on GenAI that generates text, although other modalities of output (such as image, video, or sound) are possible. Large language models (LLMs) are currently the core technological component of such systems. They are based on very large multilayered (deep) neural networks, pre-trained on vast collections of texts. A technology primer on GenAI (aimed at lawyers) can be found, for example, in Melham (2025), Frankenreiter and Livermore (2025), or Surden (2024).

The law provides a very suitable application domain for LLMs, with vast and possibly disruptive implications for legal professions and institutions. Generative AI has indeed spurred an impressive acceleration in AI and law research and applications. LLMs have become a major research focus and LLM-based systems are widely integrated into legal practice, where they can be used for multiple purposes, such as summarization, translation, drafting, reviews of literature and cases, search and extraction of information, and interactive examination of legal and factual matters (see more in Katz et al. 2023; Sayooj Devadas and Sangwan 2025; Ariai et al. 2025; D’Alotto et al. 2025; Galli and Contissa 2025). As the deployment of LLMs in legal practice has just started, it is still early to assess their impact on legal practices, though we may venture to say that their massive deployment will be relevant for legal semiotics (Wagner and Marusek 2023), by affecting the way in which we approach different determination of meaning, as they take place in different legally relevant activities, such as legal translation, the analysis of factual descriptions (e.g., expert opinions) and the interpretation of legal sources.

Progress has been rapid, as LLMs grow in size, achieve human-like or better performance on tasks requiring reasoning skills and other cognitive capacities, such as the ability to process multimedia content. It appears that they have passed both the Turing test and a growing range of qualifying exams for professional lawyers. To be sure, LLMs are not the only technology that powers GenAI tools, but it is the LLM component that has unlocked the current AI and law boom, including a new wave of thinking about AI-assisted legal interpretation.

For instance, the user can paste a legal text (e.g., a contractual clause, a statutory provision, or a judicial pronouncement) into the chat window and ask the chatbot to, for example, (1) segment provisions into operative phrases; (2) generate alternative paraphrases; (3) list or discuss ambiguities; (4) map provisions within their normative environment (other laws in the same matter, definitions, sections, recitals/purposes); and (5) articulate competing interpretive arguments tied to recognized canons of interpretation. Such outputs—the answers of the chatbot—could be valuable as they will in many cases semantically resemble what a competent person would say in response to those interpretive tasks. Here is an example from a real conversation with a GenAI chatbot (ChatGPT 5.2 Thinking) about the “No vehicles in the park” rule.

**USER INPUT:** How shall I interpret the word "vehicle" in the rule "No vehicles in the park"? Does this prohibition also include bicycles? And what about children’s bikes?

**CHATBOT OUTPUT:** “Vehicle” is a *fuzzy* word: in everyday speech it often means “car/truck/motorbike,” but in rules and ordinances it can mean “anything used to transport people or goods,” which can include bicycles.

How to interpret “**No vehicles in the park**” depends on context.

**1) Start with the most important clue: what problem is the rule trying to prevent?**

Different purposes point to different meanings:

**If the purpose is protecting pedestrians / reducing collisions / keeping paths calm**

Then “vehicle” is often read broadly → **bicycles are included** […]

**If the purpose is protecting grass/grounds or avoiding heavy damage**

Then “vehicle” is usually “things with wheels that can rut or tear up turf” → **bikes may still be included**, [...]

**If the purpose is noise, emissions, speeding, or engine danger**

Then “vehicle” is commonly shorthand for **motor vehicles** → bicycles are *less likely* to be intended [...]

**2) Legal/official meaning: many rulebooks treat bicycles as “vehicles”**

In lots of traffic-code style definitions, a bicycle is literally a “vehicle.” [...]

**3) Practical meaning: signage and layout usually reveal the intended scope** [...]

**So—does it include bicycles? Often yes, but not always.** [...]

**What about children’s bikes?**

A child’s bike is still a bicycle, so **it doesn’t magically stop being a “vehicle.”** But in practice:

- Many parks *informally* tolerate **small kids riding** at low speed, especially off busy paths. [...]

**A sensible rule of thumb**

- If you want the “safe, most defensible” reading: **assume bicycles (including kids’ bikes) are prohibited** unless there’s an explicit allowance.
- If you want the “most likely intent” reading: it depends—many parks mean “no cars/motorbikes,” but they *should* say “no motor vehicles” if that’s what they mean. [...]

*Fig. 3 A verbatim record (with some omissions) of a conversation with GenAI (ChatGPT 5.2 Thinking)*

As can be seen, the chatbot's output in Fig. 3 appears immediately useful, and its generation only required a very simple prompt. Still, it has important limitations. It is (so far) problematic to say that GenAI tools, including chatbots, would systematically ground their answers in factual "knowledge," although various techniques can guide the LLM to provide factually better answers and to pragmatically overcome this limitation (e.g., retrieval-augmented generation which supplements the user query (prompt) with links to pertinent sources). It is also problematic to say that GenAI tools would "reason" in the relevant sense, although various techniques can leverage argumentative and planning strategies to pragmatically overcome this limitation (e.g., CoT (chain-of-thought) which generates chains of reasoning steps supporting the presented claims). And it is problematic to say that GenAI tools would ground their answers in some inner "understanding" of the user prompt, although it may be argued that GenAI pragmatically manipulates the approximated "meaning" of the text via the embedded representation of vector tokens.

Importantly, LLMs' "competence" is primarily linguistic and discursive, not knowledge-based or rules-based. This is what sets it apart from the previous wave of AI and law techniques. As such, the current limits of the LLMs in the legal domain should not be underestimated: they have no sense of truth, no direct access to the physical, human, and social world; their ability to engage in reasoning and argumentation is limited and brittle; their internal working is opaque, and their explanations are often superficial or misleading; and they may provide false or unverified information (called "hallucinations"). Unthoughtful use of LLMs may thus induce legal decision-makers to adopt suggestions based on false facts, fallacious reasoning, or disregard of the human interest and the moral and legal principles at stake. It may deskill legal professionals, who may passively endorse the machine's outputs rather than submit them to rigorous factual, legal, and logical checks, and who may fail to complement such outcomes with their own creativity. It may lead citizens to act upon mistaken legal indications and promote views (possibly outdated or overruled) which happen to be well represented in the training data, rather than views that are correct by any applicable legal standards.

This does not exclude that a jurist can profitably entrust an LLM with legal tasks and specifically request an LLM to elicit interpretive issues and suggest how to address them. The LLM can, for example, quickly re-express a provision in different language registers (plain language, technical paraphrase), surface implicit structures (conditions, exceptions, definitions, cross-references), and propose candidate readings (extensive/restrictive) together with reasoning patterns and the textual cues that might support them. The jurist, however, should not assume that the LLM shares the cognitive skills that are deployed by humans in interpretation, that it possesses a critical understanding of legal and social facts, norms and values, and may accordingly apply interpretive canons. The LLM rather relies on the linguistic patterns it has learned from the large corpus of training data on which it has been trained (which may or may not include, only or mostly, good representations of canonical approaches to interpretation). Given an input prompt (the user question), the LLM has the capacity to use such patterns to generate a continuation that is statistically likely (it thus captures legal, social, and human realities only to the extent that they are reflected in linguistic patterns).

## The Importance of Data and Prompts

For real-life interpretation tasks, the interpreter needs to be sure the AI system has access to relevant data. In this regard, it is important to note a key distinction between specialized GenAI tools for lawyers and GenAI tools which are publicly available, including those embedded in popular web search engines. The specialized systems would typically utilize domain-specific data contained in some external knowledge base curated by the system providers.

This is particularly important because much of the relevant legal data is not contained in the documents used in the pre-training the LLM (legal sources, cases, or doctrinal papers and books accessible to the system at the time of the pre-training) and neither is it freely available on the Internet. Some relevant data will thus not be in the pre-training dataset behind the LLM, and neither will it be possible for LLM-based tools to retrieve relevant legal authorities from the web in the post-training stage. For textbook examples like "No vehicles in the park," this data limitation might not pose a particular difficulty, but interpretive questions about highly technical (or very recent) pieces of legislation could see the interpreter (here an AI system) to "hallucinate" an inexistent authority because a reliance on *some* authority could be the linguistic pattern (schema) learned by the LLM for this kind of interpretive task.

The specialized GenAI tools for lawyers typically seek to plug this data/knowledge gap by RAG (retrieval-augmented generation). RAG is an architecture for optimizing the performance of an artificial intelligence (AI) model by connecting it with an external knowledge base. The data in the external knowledge base is retrieved and automatically added to the user query (prompt), meaning that the LLM would then "see" all the relevant information as part of the prompt and the generated answer could thus refer specifically to the accurate information provided.

As should be clear by now, the quality of the GenAI answer (the generated output) is in large part dependent on the quality of a user prompt. RAG is one example of a prompt enhancement technique. A system prompt is another technique. The quality of the output can also be enhanced by user-led prompting techniques, whereby the user ensures their input contains a full set of relevant variables that will help steer the model answer into the right direction (Meincke et al. 2025). Various studies develop prompting techniques for legal tasks specifically (Yu et al. 2022; Müller-Peltzer et al. 2025) and private vendors include many more in their standardized prompt libraries for their paying customers. Importantly, this ability of the user to somehow influence what GenAI "sees" and "does" via strategically crafted prompts has motivated a series of experiments in what is now called "generative" legal interpretation.

## Generative Interpretation

Generative interpretation is a term coined by Hoffman and Arbel (2024) as part of their proposal to use LLMs as a practical aid for contract interpretation. They argue that courts—instead of relying mainly on dictionaries and canons (textualism) or open-ended testimony and context (contextualism)—can use LLMs to find what the parties meant at the time of contracting by analyzing the contract's language and relevant extrinsic evidence in a disciplined, replicable way. The techniques and tools deployed for such generative interpretation are existing publicly available LLMs and structured prompts.

The justification for this approach is that LLMs can process large amounts of text and are sensitive to linguistic context. As such, LLMs can help courts estimate ordinary meaning "in context" of other relevant documents used by the parties to the contract. LLMs can help quantify how ambiguous a term is, fill contractual gaps, and even evaluate the marginal probative value of particular pieces of extrinsic evidence—while keeping costs down and reducing the ad hoc feel of current court practice in contract interpretation cases. See also Arbel (2025, 2026), Calvert (2025). Generative interpretation can be seen as part of a broader empirical turn in jurisprudence (Bystranowski et al. 2025), in the attempt to naturalize it (Lee 2024).

A similar idea has been around since around 2020, when scholars discussed how computer-assisted methods could be used to ascertain "ordinary meaning" for the purposes of textual interpretation of legal sources, including statutes. At that time, the focus was on the methods of corpus linguistics (Lee and Mouritsen 2018; Tobia 2020; Nyarko and Sanga 2022), while now the same idea is tested via GenAI tools based on LLMs (Engel and McAdams 2024). Judges have been experimenting with the generative interpretation approach in their work too (Miller 2025), with some scholars noting a particular advantage of this approach for US constitutional interpretation (a textualist approach: Coan and Surden 2025) or statutory simplification (Hariri and Ho 2026). There are attempts to justify the use of generative interpretation by comparing this method's outputs with reliance on dictionaries and corpus linguistics (Johnson 2025).

Because of the LLMs' linguistic competence, it is not easy to draw a hard line between generative interpretation *stricto sensu* and other examples of legal tasks that require LLMs to interpret legal documents. This concerns, for example, summarization of legal documents (Dal Pont et al. 2023; Deroy et al. 2025; Santosh et al. 2024; Benedetto et al. 2025; Zin et al. 2025), assessment of the quality of legal documents (Boi et al. 2025), or classifications of legal interpretation methods in existing legal document by using LLMs (Dugac and Altwicker 2025). The technical steps involved in the production of these outputs will be the same, given that GenAI is a general-purpose technology. Vast opportunities exist for the use of LLMs for the purpose of knowledge extraction, which may include the translation of legal sources into knowledge bases, to be processed according to logic or argumentation (Dal Pont et al. 2026). Some projects exist that aim at training LLMs in being proficient in legal argumentation, by prompting them to present arguments that comply with legal methods, e.g., with interpretation canons.

The ability of LLMs to competently process natural language input and generate seemingly competent language output has also led to a renowned interest in Agentic AI systems, including multi-agent systems that have access to (and can autonomously control) a range of other software tools, including rules-based and knowledge-based tools. This makes the field of enquiry even more complex, and it is not always easy to say whether an interpretive task has been carried out by an LLM as such. Rather, it is more precise today to conceptualize these tasks as LLM-enabled interpretations and to explore precisely the methodical and social legitimacy of specific LLM-enabled use cases, taking into account the back-end architecture of the AI system and the data it utilizes.

## Evaluation of Outputs

A specific problem of LLMs in the legal domain pertains to the evaluation of their outputs. No precisely defined ground-truth (factual data) exists for checking, for instance, whether the LLM's answer about the "No vehicles in the park" rule is correct or incorrect, justified, or unjustified. The performance of LLM-enabled tasks in law is thus measured by benchmarks (Guha et al. 2025; Chalkidis et al. 2022). Benchmarking is about how well an LLM performs on a legal task X, whereby "well" may be assessed by a human-led jury and compared to human-produced outputs for the same task. In interpretive tasks, such benchmarks can be formed by authoritative precedent decisions that confirm (or deny) certain interpretations of the source document, but it could also be a human-led scoring based on how persuasive the interpretive outputs (and supporting arguments) are for a human jury (e.g., Garzo and Palumbo 2025). Legal analysis tasks are thus usually assessed using qualitative benchmarks (Choi and Schwarcz 2025), rather than quantitative ones.

LLMs generative powers can be leveraged for other kinds of tasks, some of which may be utilized in the automated workflow that is motivated by an interpretive task. For example, LLMs can be used to automate annotation of documents (Libal et al. 2026), retrieve and generate legislative definitions (Zilli et al. 2025), or detect unfair clauses in terms of service (Panarelli et al. 2026). In such tasks, their performance may be measured by traditional quantitative metrics, but it may come at the cost of reproducibility and consistency of the answers, which are notorious pain points of the non-deterministically behaving GenAI systems (Cohen-Sasso 2025; Floridi 2025a).

Among the works on evaluating and benchmarking the results obtained by LLMs, see, e.g., Fei et al. (2024), Tan et al. (2024), Chlapanis et al. (2024), Vals AI (2025).

## Critiques of LLM-Enabled Interpretation

The key normative-practical question concerning the use of LLMs in legal interpretation pertains to the extent to which they can play a useful role, contributing not only to efficiency but also to the quality of legal analysis. A number of criticisms raised against the use of LLMs are relevant in this regard.

It has been argued that the GenAI advantages (such as significant time savings in contract review, Martin et al. 2024) may be offset by the professional imperative to verify any LLM-generated output (Yuvaraj 2025), as stated by many professional guidelines. Such controls are necessitated by reported rates of hallucinations (17–34% on benchmarked legal queries according to Magesh et al. 2025). Unfortunately, hallucinations are an intrinsic feature of GenAI systems and cannot be eliminated by LLMs alone (Karpowicz 2025), though they can be limited and controlled, e.g., by requiring the GenAI tool to look on the Internet for all the sources it cites.

The cost of mitigating LLMs' hallucination in law is high (Schepers et al. 2025) and better-performing LLMs may not solve the verification-value tension in the context of the law. It has been argued that legal interpretation tasks require valid (not better or worse) performance (Hildebrandt 2025), though human interpretation may also fail to meet correctness standards.

While RAG may be seen as a technique to mitigate hallucinations and anchor generated text in good legal sources, it has important limitations. The GenAI tool needs to know which documents to retrieve, and for this it needs to decide which documents/chunks are relevant for the task. The relevance is based on a compressed external knowledge database and the retrieval process thus presents a trade-off between speed, relevance, and completeness of retrieved parts of the database (Ferraris et al. 2024; Wiggers and Hartz 2024; Weller et al. 2025). Further research is needed to establish in what contexts and to what extent RAG may effectively reduce hallucinations.

A cognate point of attack is consistency: it has been observed that on the one hand LLMs may provide different responses given the same prompt and on the other hand that their responses may be different from the responses given by humans. It has also been observed that such inconsistencies have distinct effects: LLMs inconsistency has a larger impact on foreseeability, while human inconsistency has a larger impact on fairness (Cohen-Sasso 2025). Others too highlight the unreliability of LLMs in high-stakes application areas such as legal interpretation and call the generated insights "artificial intuition," rather than empirical analysis of any meaning (Lee and Egbert 2024).

It has been observed that statistic-based technologies such as LLMs fail to engage with legal sources at a jurisprudential (normative) level of analysis and instead only process the legal data as facts, limiting their ability to produce jurisprudential insights that would be readily informative for a participant in the legal system, as opposed to insight that would be informative for an external observer of the system (Janeček 2023). It has also been observed that they fail to deliver the ordinary meaning of legislative sources, namely, what an ordinary reader would think the source means (Chen 2025). This is related to the fact that they do not share our understanding of natural language. Cognate themes can be found in the literature focusing on distant reading and distant writing (Floridi 2025b), which well captures the tensions between the ability to read and the ability to write, when conceived as a data-driven ability.

The (un)reliability of LLMs in interpretive tasks has been assessed through various empirical tests. For example, studies show that LLMs incur in significant failures when extracting principles of law (Molinari et al. 2025); detecting unfair clauses in terms of service (Panarelli et al. 2026); and reasoning with facts and statutes (Blair-Stanek et al. 2023), including if they try to follow the IRAC structure (Linna and Linna 2025).

Some critics have pointed to possible misuses of LLMs in legal interpretation. For instance, Grimmelmann et al. (2026) observed that the "[t]he superficial fluency of LLM-generated text conceals fundamental gaps between what these models are currently capable of and what legal interpretation requires to be methodologically and socially legitimate." According to these authors, generative interpretation methods are not demonstrably reliable (reproducible and consistent), and they have an epistemic gap (it is not clear that what they do is what we think they do). Waldon et al. (2025) have argued that legal decision makers should not take interpretive suggestions of LLMs as authoritative indications of correct interpretation, but rather as "brainstorming" tools, to expand the set of interpretive suggestions and arguments that the judge might consider.

In light of the criticisms just presented, we may conclude that LLMs should never be used as authoritative oracles providing reliable legal advice on interpretation, even when their interpretive proposals are accompanied by apparently sound legal arguments. However, their ability to organize, present, and contextualize language-embedded knowledge can support their use as companions for skilled legal interpreters: LLMs may synthesize at least some doctrinal debates on the interpretive issues at stake and suggest possible interpretations as well as indicate reasons for and against their acceptance. The lawyers using such companions, however, should maintain direct contact with the original legal sources, continue to use their trained expertise in figuring and solving interpretive issues, and carefully check the correctness of all arguments and references proposed by the LLMs. They should have the final decision, and full responsibility, on what interpretation to endorse based on what arguments: they should not accept an interpretation just because it was proposed by an LLM; acceptance should be grounded on a normative justification they endorse (though the LLM may support them in finding such a justification).

This is not an optional recommendation that would be based on a comparative strength of human interpreters relative to LLMs. Rather, it is a recommendation based on LLMs' cognitive and technical inability to discharge these significant legal tasks autonomously. It is also why it seems to us that the delegation to LLMs interpretive tasks in the context of agentic AI should be considered with an extreme caution, since in such a context the LLMs would take action based on the interpretation it has adopted, without direct human control. Such an approach would directly incur all the criticisms we have just presented. Little Tom would probably object to being fined by some LLM-enabled system since the LLM inside the machine has autonomously (without a human-led validation) interpreted the "No vehicles in the park" rule as including children's bikes.

This brings us back to the initial criticism that the use of LLMs in legal interpretation may not be justified under a cost-benefit analysis, as the cost of checking the LLMs suggestions may outweigh the benefit of having such suggestions. Whether this is the case depends on the nature of the interpretive problems being considered (e.g., an LLM is likely to be most useful where a vast doctrinal debate exists and where that debate is well articulated in digitized literary sources). It also depends on the traditional information tools already available to the lawyer (a textbook or commentary may provide lawyers with all the information they need) and on the lawyer's capacity to question the LLM to extract information that is relevant to the interpretive tasks. We may however venture to suggest that as the LLMs' competence and "learned" knowledge base grow, so would their deployment in legal tasks. Currently, the reasonable expectation is that the use of LLMs in support of legal interpretation tasks would become less error-prone, for example, through an additional layer of automated techniques (e.g., post hoc legal citation verifications) or through smart practical implementations into robust workflows (e.g., LLM-enabled detection of inconsistencies or interpretive mistakes in a source document). This development could make verification less costly or it could side step the verification-value paradox entirely.

# 6. Conclusion

The times have not yet arrived where little Tom could rely on an AI system to know whether after all he can enter the park with his small bike, on the basis of the system’s interpretation of the “No vehicles in the park” rule. The human cognitive skills required for legal interpretation are not currently possessed by AI tools, and it is unlikely that they could be obtained through the incremental development of existing technologies.

However, there is ample space for the deployment of AI as supporting tools for lawyers engaged in interpretive reasoning. In this vein, AI’s interpretive suggestions and arguments should be viewed as additional information provided to a competent lawyer rather than as independently reliable legal opinions. Effort should be put in supporting lawyers who wish to use LLMs for the purpose of legal interpretation. At present, an appropriate methodical guidance on how to use AI systems (rather a guidance on how not to use them) is missing.

Interesting lines of research pertain to integrating approaches based on logic and argumentation with LLMs. One possibility consists in using the power of LLMs for scalable extraction of information from legal documents or documents that evidence legally relevant behaviours and the translation of the extracted information into computable formalisms, in order to obtain a knowledge base that can support logical reasoning and argumentation (including in addressing interpretive issues). Another possibility pertains to teaching LLMs to address interpretive issues, by prompting or training them to correctly apply interpretive canons and legal theories. A further possibility pertains to controlling the interpretive outcomes by LLMs by relying on logic, formal argumentation, and other standards for legal reasoning (such controls may also be in part delegated to other LLMs systems, prompted-trained to execute such controls). And yet another possibility is in (re)designing the lawyer’s workflows such that the benefits of different AI systems can be fully leveraged and that the risks discussed in this chapter are properly mitigated.